%% file: main.tex
\documentclass{article} 
\usepackage{arxiv,times}
\usepackage{natbib}
\DeclareUnicodeCharacter{0301}{\'{e}}

\input{math_commands.tex}

\usepackage{url}
\usepackage{hyperref}
\usepackage[american]{babel}
\usepackage{microtype}
\usepackage{graphicx}
\usepackage{subcaption}
\usepackage{tabularx}
\usepackage{multirow}
\usepackage{paralist}
\usepackage{bbm}
\usepackage{mathtools}
\usepackage{booktabs}
\input{notation}

\usepackage{pifont}
\usepackage{tikz}
\newcommand{\tikzcircle}[2][red,fill=red]{\tikz[baseline=-0.5ex]\draw[#1,radius=#2] (0,0) circle ;}%

\newcommand{\xhdr}[1]{\vspace{0em}\noindent{{\bf #1.}}}

\usepackage{colortbl}

\definecolor{Gray}{gray}{0.9}
\definecolor{LightCyan}{rgb}{0.88,1,1}
\newcolumntype{a}{>{\columncolor{Gray}}c}
\newcolumntype{b}{>{\columncolor{white}}c}

\title{Rethinking Stability for Attribution-based Explanations}

\author{Chirag~Agarwal$^{1}$, Nari~Johnson$^{2}$, Martin~Pawelczyk$^{3}$, Satyapriya~Krishna$^{4}$,\\ \textbf{Eshika~Saxena}$^{4}$, \textbf{Marinka~Zitnik}$^{4}$ \& \textbf{Himabindu~Lakkaraju}$^{4}$\\
$^{1}$ Media and Data Science Research Lab, Adobe \\
$^{2}$ Carnegie Mellon University \\
$^{3}$ University of Tübingen \\
$^{4}$ Harvard University \\
}

\newcommand{\hide}[1]{}

\begin{document}

\maketitle

\begin{abstract}
\input{000abstract}
\end{abstract}

\section{Introduction}
\label{sec:intro}
\input{010intro}

\section{Related Works}
\label{sec:related}
\input{020related}

\section{Stability Analysis for Evaluating Explanations}
\label{sec:stability}
\input{030metric}

\section{Experiments}
\label{sec:expt}
\input{040expt}

\section{Conclusion}
\label{sec:concl}
\input{050concl}



\clearpage
\bibliography{main}

\newpage
\appendix
\input{060appendix.tex}

\end{document}

%% file: math_commands.tex
\usepackage{amsmath,amsfonts,bm}

\def\eqref#1{equation~\ref{#1}}

\def\1{\bm{1}}

\DeclareMathAlphabet{\mathsfit}{\encodingdefault}{\sfdefault}{m}{sl}
\SetMathAlphabet{\mathsfit}{bold}{\encodingdefault}{\sfdefault}{bx}{n}

\DeclareMathOperator*{\argmax}{arg\,max}

%% file: notation.tex
\def\to{{\,\rightarrow\,}}

\mathchardef\mhyphen="2D

\newcommand{\vertiii}[1]{{\left\vert\kern-0.25ex\left\vert\kern-0.25ex\left\vert #1
    \right\vert\kern-0.25ex\right\vert\kern-0.25ex\right\vert}}

\newcommand{\bX}[0]{\mathbf{X}}

\def\bm{{\mathbf{m}}}

\def\bx{{\mathbf{x}}}
\def\by{{\mathbf{y}}}

\def\bX{{\mathbf{X}}}

\def\cE{\mathcal{E}}

\def\cL{\mathcal{L}}

\def\cX{\mathcal{X}}
\def\cY{\mathcal{Y}}



%% file: 000abstract.tex
As attribution-based explanation methods are increasingly used to establish model trustworthiness in high-stakes situations, it is critical to ensure that these explanations are stable, e.g., robust to infinitesimal perturbations to an input. However, previous works have shown that state-of-the-art explanation methods generate unstable explanations. Here, we introduce metrics to quantify the stability of an explanation and show that several popular explanation methods are unstable. In particular, we propose new \textit{Relative Stability} metrics that measure the change in output explanation with respect to change in input, model representation, or output of the underlying predictor. Finally, our experimental evaluation with three real-world datasets demonstrates interesting insights for seven explanation methods and different stability metrics.

%% file: 010intro.tex
With machine learning (ML) models being increasingly employed in high-stakes domain such as criminal justice, finance, and healthcare, it is essential to ensure that the relevant stakeholders understand these models' decisions. 
\hide{However, state-of-the-art ML models are complex black-box models and hard highly complex and therefore to interpret. To address this problem, several approaches have been proposed in recent literature to explain model decisions using simplifications of models, feature sub-selection, or attention~\citep{ross2017right,kim2018interpretability,carter2019made}. In this work, we focus on attribution-based explanation methods~\citep{simonyan2013saliency,smilkov2017smoothgrad,sundararajan2017axiomatic,ribeiro16:kdd,lundberg2017unified} that have recently received significant research attention.  However, existing explanation methods suffer from different drawbacks.}
However, existing approaches to explain the predictions of complex machine learning (ML) models suffer from several critical shortcomings. Recent works have shown that explanations generated using attribution-based methods are not stable~\citep{ghorbani2019interpretation,slack2019can,dombrowski2019explanations,adebayo2018sanity,alvarez2018robustness,bansal2020sam}, e.g. that infinitesimal perturbations to an input can result in substantially different explanations. 

Existing metrics~\citep{alvarez2018robustness} measure the change in explanation only with respect to the input perturbations, e.g., they only assume black-box access to the predictive model, and don't leverage potentially meaningful information such as the model's internal representations to evaluate stability. To address these limitations of existing stability metrics, we propose \textit{Relative Stability} that measures the change in output explanation with respect to the behavior of the underlying predictive model~(Section~\ref{sec:rel-stab}). Finally, we present extensive theoretical and empirical analysis~(Section~\ref{sec:result}) for comparing the stability of seven state-of-the-art explanation methods using multiple real-world datasets.


%% file: 020related.tex
This paper draws from two main areas of prior work: 1) attribution-based explanation methods, and 2) stability analysis of explanations.

\xhdr{Attribution-based Explanation Methods} While a variety of approaches have been proposed to explain model decisions for classifiers, our work focuses on \emph{local feature attribution explanations}, which measure the contribution of each feature to the model's prediction on a point. In particular, we study two broad types of feature attribution explanations: gradient-based and approximation-based. Gradient-based feature attribution methods like VanillaGrad~\citep{simonyan2013saliency}, SmoothGrad~\citep{smilkov2017smoothgrad}, Integrated Gradients~\citep{sundararajan2017axiomatic}, and Gradient$\times$Input~\citep{shrikumar2017learning} leverage model gradients to quantify how a change in each feature would affect the model's prediction. Approximation-based methods like LIME~\citep{ribeiro2016should}, SHAP~\citep{lundberg2017unified}, Anchors~\citep{ribeiro2018anchors}, BayesLIME, and BayesSHAP~\citep{slack2021reliable} leverage perturbations of individual inputs to construct a local approximation model from which feature attributions are derived.

\hide{\njedit{While a variety of approaches have been proposed to explain model decisions for classifiers, our work focuses on \emph{local feature attribution explanations}, which measure the contribution of each feature to the model's prediction at a point.  Specifically, we study two broad types of feature attribution explanations: gradient based and approximation based.} \njdelete{For instance,} \njedit{Gradient based feature attribution methods} methods like VanillaGrad~\citep{simonyan2013saliency}, SmoothGrad~\citep{smilkov2017smoothgrad}, Integrated Gradients~\citep{sundararajan2017axiomatic}, and Gradient$\times$Input~\citep{shrikumar2017learning}, \njdelete{are gradient-based local explanations as they leverage gradients computed with respect to input dimensions of individual instances} \njedit{leverage model gradients with respect to a point to quantify how a change in each feature would affect the model's prediction. } \njedit{Approximation based methods like}\njdelete{On the other hand, methods like} LIME~\citep{ribeiro2016should}, SHAP~\citep{lundberg2017unified}, Anchors~\citep{ribeiro2018anchors}, BayesLIME, and BayesSHAP~\citep{slack2021reliable} leverage perturbations of individual inputs to construct \njdelete{interpretable local approximations.} \njedit{a local approximation model from which feature attributions are derived.}}

\looseness=-1
\xhdr{Explanation Stability} Recent works have formalized desirable properties for feature attribution explanations~\citep{agarwal2022probing}. Our work specifically focuses on the \emph{stability} of explanations. \citet{alvarez2018robustness} argued that ``similar inputs should lead to similar explanations'' and is the first work to formalize a metric to measure the stability of local explanation methods. We highlight potential issues with this stability metric that measures stability only \textit{w.r.t.} the change in \emph{input}.

\hide{Recent works have highlighted the need for several critical desirable properties in explanation methods~\citep{agarwal2022probing}. In particular, these methods are highly sensitive to small changes in inputs~\citep{ghorbani2019interpretation,bansal2020sam} and unfaithful to the underlying machine learning model~\citep{sippy2020data}. Attribution-based explanation techniques like LIME and SHAP generate explanations that vary between different runs of the algorithms~\citep{zhang2019should,lee2019developing,bansal2020sam}, and hyperparameters used to select the perturbations can influence the output explanation~\citep{zhang2019should,bansal2020sam}. Finally, some prior works have attempted to both quantify~\citep{alvarez2018robustness} and tackle~\citep{yeh2019fidelity} the problem of stability of explanations.}

%% file: 030metric.tex
\subsection{Notation and Preliminaries}
\label{sec:notation}
\xhdr{Machine Learning Model} Given a feature domain $\cX$ and label domain $\cY$, we denote a classification model $f{:}~\cX{\to}\cY$ that maps a set of features $\bx {\in} \cX$ to labels $\by {\in} \cY$, where $\bx\in\mathbb{R}^{d}$ is a $d$-dimensional feature vector, $\by\in\{0, 1, \dots, \text{C} \}$ where C is the total number of classes in the dataset. We use $\bX=\{\bx_{1}, \bx_{2}, \dots, \bx_{N}\}$ to denote all the $N$ instances in the dataset. In addition, we define $f(\bx){=}\sigma(h(\bx))$, where $h: \cX{\to}\mathbb{R}$ is a scoring function (e.g., logits) and $\sigma: \mathbb{R}{\to}\cY$ is an activation function that maps output logit scores to discrete labels. Finally, for a given input $\bx$, the output predicted class label is: $\hat{y}_{\bx}{=}\argmax_c f(\bx)$. We assume access to the gradients and intermediate representations of model $f$.

\xhdr{Explainability Methods} An attribution-based explanation method $\cE$ generates an explanation $\mathbf{e}_{\bx}\in\mathbb{R}^{d}$ to explain model prediction $f(\bx)$. 
To calculate our stability metrics, we generate perturbations $\bx'$ by adding infinitesimal noise to $\bx$, and denote their respective explanation as $\mathbf{e}_{\bx'}$. 

\subsection{Existing Definition and Problems}
\label{sec:existing}
\hide{Formally, stability refers to the robustness of explanations -- \textit{similar inputs} should give rise to \textit{similar explanations}. For a given instance $\bx$, a similar or perturbed instance $\bx'$ is generated by adding \textit{infinitesimally small noise} to the clean instance $\bx$ such that $\hat{y}_{\bx} = \hat{y}_{\bx'}$. Alvarez et al.~\citep{alvarez2018robustness} defines a local notion of stability, i.e., for neighboring inputs where it relies on a point-wise neighborhood-based local Lipschitz continuity:}

\citet{alvarez2018robustness} formalize the first stability metric for local explanation methods, arguing that explanations should be robust to local perturbations of an input. To evaluate the stability of an explanation for instance $\mathbf{x}$, perturbed instances $\mathbf{x'}$ are generated by adding infinitesimally small noise to the clean instance $\mathbf{x}$ such that $\hat{y}_{\mathbf{x}} = \hat{y}_{\mathbf{x'}}$:
\begin{align}
    & \text{S}(\mathbf{x},\mathbf{x'},\mathbf{e}_{\mathbf{x}},\mathbf{e}_{\mathbf{x'}}) = \max_{\mathbf{x'}} \frac{||~ \mathbf{e}_{\mathbf{x}}-\mathbf{e}_{\mathbf{x'}}~||}{||~\mathbf{x}-\mathbf{x'}~||},~\forall \mathbf{x'}~\text{s.t.}~\mathbf{x'}\in\mathcal{N}_{\mathbf{x}};~\hat{y}_{\mathbf{x}}=\hat{y}_{\mathbf{x'}}
    \label{eq:old_stab1}
\end{align}
where $\mathcal{N}_{\mathbf{x}}$ is a neighborhood of instances $\mathbf{x}'$ similar to $\mathbf{x}$, and $\mathbf{e}_{\mathbf{x}}$ and $\mathbf{e}_{\mathbf{x'}}$ denote the explanations corresponding to instances $\mathbf{x}$ and  $\mathbf{x}'$, respectively. For each point $\bx'$, the stability ratio in Equation~\ref{eq:old_stab1} measures how the output explanation varies with respect to the change in the \emph{input}. Because the neighborhood of instances $\mathcal{N}_{\mathbf{x}}$ are sampled to be similar to the original instance $\mathbf{x}$, the authors argue that points that are similar should have similar model explanations, e.g., we desire the ratio in Equation~\ref{eq:old_stab1} to be close to $1$~\citep{alvarez2021from}. This stability definition relies on the point-wise neighborhood-based local Lipschitz continuity of the explanation method $\mathbf{e}_{\bx}$ around $\mathbf{x}$.

\begin{figure}[h]
    \centering
    \vspace{-2mm}
    \includegraphics[width=0.99\textwidth]{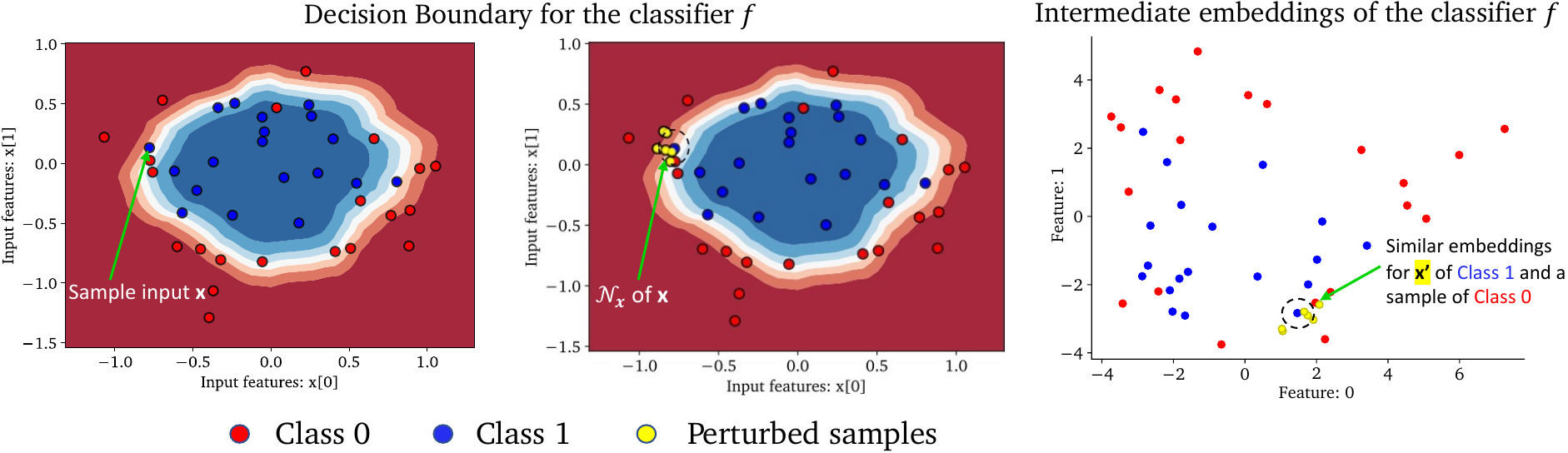}
    \vspace{-2mm}
	\caption{Decision boundaries and embeddings of a two-layer neural network predictor $f$ with 100 units trained on the \href{https://scikit-learn.org/stable/modules/generated/sklearn.datasets.make_circles.html}{circles} dataset. The heatmaps (left and middle column) shows the models' confidence for the positive-class (in blue), test set examples $\bx$ (\tikzcircle{2pt}, \tikzcircle[blue,fill=blue]{2pt}), and a set of perturbed samples $\bx'$ (\tikzcircle[yellow,fill=yellow]{2pt}). While all perturbed samples $\mathbf{x}'$ are predicted to the same class as $\mathbf{x}'$, the embeddings (right column) for some $\mathbf{x}'$ are far from the embeddings of $\mathbf{x}'$ and similar to the embeddings of Class 0, highlighting the need of incorporating the model behavior using its internal embeddings (Equations~\ref{eq:rrs},\ref{eq:ros}).}
	\vspace{-2.5mm}
    \label{fig:toy}
\end{figure}

\xhdr{Problems} We note two key problems with the existing stability definition: i) it only assumes black-box access to the prediction model $f$, and does not leverage potentially meaningful information such as the model's internal representations for evaluating stability; and ii) it implicitly assumes that $f$ has the same \emph{behavior} on inputs $\bx$ and $\bx'$ that are similar.  While this may be the case for underlying prediction models that are smooth or robust, this assumption may not hold in a large number of cases. In Figure~\ref{fig:toy}, we discuss a toy example where perturbed samples $\mathcal{N}_{\bx}$ have drastically different intermediate representations than the original point $\bx$.
Note that since the goal of an explanation is to faithfully and accurately represent the behavior of the underlying prediction model~\citep{agarwal2022probing}, we argue that an explanation method \emph{should} vary for points $\bx$ and $\bx'$ where the prediction model's behavior differs. Thus, we argue for the inclusion of new stability metrics that measure how much explanations vary with respect to the behavior of the underlying prediction model.


\subsection{Proposed metric: Relative Stability}
\label{sec:rel-stab}
To address the aforementioned challenges, we propose \textit{Relative Stability} that leverages model information to evaluate the stability of an explanation with respect to the change in the a) input data, b) intermediate representations, and c) output logits of the underlying prediction model.

\xhdr{a) Relative Input Stability} We extend the stability metric in Equation~\ref{eq:old_stab1} and define \textit{Relative Input Stability} that measures the relative distance between explanations $\mathbf{e}_{\bx}$ and $\mathbf{e}_{\bx'}$ with respect to the distance between the two inputs $\bx$ and $\bx'$.
\begin{align}
\text{RIS}(\bx, \bx', \mathbf{e}_{\bx}, \mathbf{e}_{\bx'}) &= \max_{\bx'}\frac{|| \frac{(\mathbf{e}_{\bx} - \mathbf{e}_{\bx'})}{\mathbf{e}_{\bx}} ||_p}{ \max(|| \frac{(\bx - \bx')}{\bx}||_p, \epsilon_{min})},~\forall \bx'~\text{s.t.}~\bx'\in\mathcal{N}_{\bx};~\hat{y}_{\bx}=\hat{y}_{\bx'}
\label{eq:ris}
\end{align}
where the numerator of the metric measures the $\ell_p$ norm of the \emph{percent change} of explanation $\mathbf{e}_{\bx'}$ on the perturbed instance $\bx'$ with respect to the explanation $\mathbf{e}_{\bx}$ on the original point $\bx$, the denominator measures the $\ell_p$ norm between (normalized) inputs $\bx$ and $\bx'$ and the $\max$ term prevents division by zero in cases when norm $|| \frac{(\bx - \bx')}{\bx}||_p$ is less than some small $\epsilon_{min}{>}0$. Here, we use the percent change from the explanation on the original point to the explanation on the perturbed instance in contrast to the absolute difference between the explanations (as in Equation~\ref{eq:old_stab1}) to enable comparison across different attribution-based explanation methods that have vastly different ranges or magnitudes. Intuitively, one can expect \textit{similar} explanations for points that are similar -- the percent change in explanations (numerator) should be \emph{small} for points that are close, or have a \emph{small} $l_p$ norm (denominator). Note that the metric in Equation~\ref{eq:ris} measures instability of an explanation and higher values indicate higher instability.

\looseness=-1
\xhdr{b) Relative Representation Stability} Previous stability definitions in Equation~\ref{eq:old_stab1}-\ref{eq:ris} do not cater to cases where the model uses different logic paths (e.g., activating different neurons in a deep neural network) to predict the same label for the original and perturbed instance. In addition, past works have presented empirical evidence that the intermediate representations of a model are related to the underlying behavior or reasoning of the model~\citep{agarwal2021towards}. Thus, we leverage the internal features or representation learned by the underlying model and propose \textit{Relative Representation Stability} as:
\begin{align}
\text{RRS}(\bx, \bx', \mathbf{e}_{\bx}, \mathbf{e}_{\bx'}) &= \max_{\bx'}\frac{|| \frac{(\mathbf{e}_{\bx} - \mathbf{e}_{\bx'})}{\mathbf{e}_{\bx}}||_p}{ \max(|| \frac{(\cL_{\bx}{-}\cL_{\bx'})}{\cL_{\bx}}||_p, \epsilon_{min})},~\forall \bx'~~\text{s.t.}~~\bx'\in\mathcal{N}_{\bx};~~\hat{y}_{\bx}{=}\hat{y}_{\bx'}
\label{eq:rrs}
\end{align}
where $\cL(\cdot)$ denotes the internal model representation, e.g., output embeddings of hidden layers, and $\delta$ is an infinitesimal constant. Due to insufficient knowledge about the data generating mechanism, we follow the perturbation mechanisms described above to generate perturbed samples $\bx'$ but use additional checks to ensure that for certain perturbations the model behaves similar to its training behavior. For any given instance $\bx$, we generate $m$ local perturbed samples such that $||\bx - \bx'||_{p} \leq \epsilon$, and $\hat{y}_{\bx}{=}\hat{y}_{\bx'}$. For every perturbed sample, we calculate the difference in their respective explanations and using Equation~\ref{eq:rrs} calculate the relative stability of an explanation. Note that, as before, the metric in Equation~\ref{eq:rrs} measures instability of an explanation and higher values indicate higher instability.

Finally, we show that the \textit{Relative Input Stability} can be bounded using the Lipschitzness of the underlying model. In particular, we proof that RIS is upper bounded by a product of the Lipschitz constant $L_{1}$ of the intermediate model layer (assuming a neural network classifier) and our proposed \textit{Relative Representation Stability}. See Appendix~\ref{sec:theory} for the complete proof.
\begin{align}
    \text{RIS}(\bx, \bx', \mathbf{e}_{\bx}, \mathbf{e}_{\bx'}) < \lambda_{1}L_{1} \times~\text{RRS}(\bx, \bx', \mathbf{e}_{\bx}, \mathbf{e}_{\bx'})
    \label{eq:proof_1}
\end{align}

\xhdr{c) Relative Output Stability} Note that Relative Representation Stability assumes that the underlying ML model is white-box, i.e., explanation method has access to the internal model knowledge. Hence, for black-box ML models we define \textit{Relative Output Stability} as:
\begin{align}
\text{ROS}(\bx, \bx', \mathbf{e}_{\bx}, \mathbf{e}_{\bx'}) &= \max_{\bx'}\frac{|| \frac{(\mathbf{e}_{\bx} - \mathbf{e}_{\bx'})}{\mathbf{e}_{\bx}} ||_p}{ \max(|| h(\bx){-}h(\bx')||_p, \epsilon_{min})},~\forall \bx'~~\text{s.t.}~~\bx'\in\mathcal{N}_{\bx};~~\hat{y}_{\bx}{=}\hat{y}_{\bx'}
\label{eq:ros}
\end{align}
where $h(\bx)$ and $h(\bx')$ are the output logits for $\bx$ and $\bx'$, respectively. Again, we proof that RRS is upper bounded by a product of the Lipschitz constant $L_{2}$ of the output model layer and our proposed \textit{Relative Output Stability}. See Appendix~\ref{sec:theory} for the complete proof.
\begin{align}
    \text{RRS}(\bx, \bx', \mathbf{e}_{\bx}, \mathbf{e}_{\bx'}) < \lambda_{2}L_{2}~\times~\text{ROS}(\bx, \bx', \mathbf{e}_{\bx}, \mathbf{e}_{\bx'})
    \label{eq:proof_3}
\end{align}

%% file: 040expt.tex
To demonstrate the utility of relative stability, we systematically compare and evaluate the stability of seven explanation methods using three real-world datasets using equations defined in Section~\ref{sec:stability}. Further, we show that, in contrast to relative input stability, relative representation and output stability better captures the stability of the underlying black-box model.

\subsection{Datasets and Experimental Setup}
\label{sec:setup}
\xhdr{Datasets} We use real-world structured datasets to empirically analyze the stability behavior of explanation methods and consider 3 benchmark datasets from high-stakes domains: i) the \textit{German Credit} dataset~\citep{Dua:2019} which has records of 1,000 clients in a German bank. The downstream task is to classify clients into good or bad credit risks, ii) the \textit{COMPAS} dataset~\citep{larson2016we} which has records of 18,876 defendants who got released on bail at the U.S state courts during 1990-2009. The dataset comprises of features representing past criminal records and demographics of the defendants and the goal is to classify them into bail or no bail, and iii) the \textit{Adult} dataset~\citep{Dua:2019} which has records of 48,842 individuals including demographic, education, employment, personal, and financial features. The downstream task is to predict whether an individual’s income exceeds \$50K per year.

\xhdr{Predictors} We train logistic regression (LR) and artificial neural network (ANN) as our predictive models. Details in Appendix \ref{apdx:model-training}. 

\xhdr{Explanation methods} We evaluate seven attribution-based explanation methods, including VanillaGrad~\citep{simonyan2013saliency}, Integrated Gradients~\citep{sundararajan2017axiomatic}, SmoothGrad~\citep{smilkov2017smoothgrad}, Input$\times$Gradients~\citep{shrikumar2017learning}, LIME~\citep{ribeiro2016should}, and SHAP~\citep{lundberg2017unified}. Following \citet{agarwal2022probing}, we also include a random assignment of importance as a control setting.  Details on implementation and hyperparameter selection for the explanation methods are in Appendix~\ref{apdx:impl-exp-methods}.

\xhdr{Setup} For each dataset and predictor, we: (1) train the prediction model on the respective dataset; (2) randomly sample $100$ points from the test set; (3) generate $50$ perturbations for each point in the test set; (4) generate explanations $\mathbf{e}_{\bx'}$ for each test set point and its perturbations using seven explanation methods; and (5) evaluate the stability of the explanations for these test points using all stability metrics~(Equations~\ref{eq:ris},\ref{eq:rrs},\ref{eq:ros}).

\begin{figure*}[t]
\centering
\vspace{-2mm}
\begin{subfigure}{0.95\linewidth}
    \centering
    \includegraphics[width=0.95\textwidth]{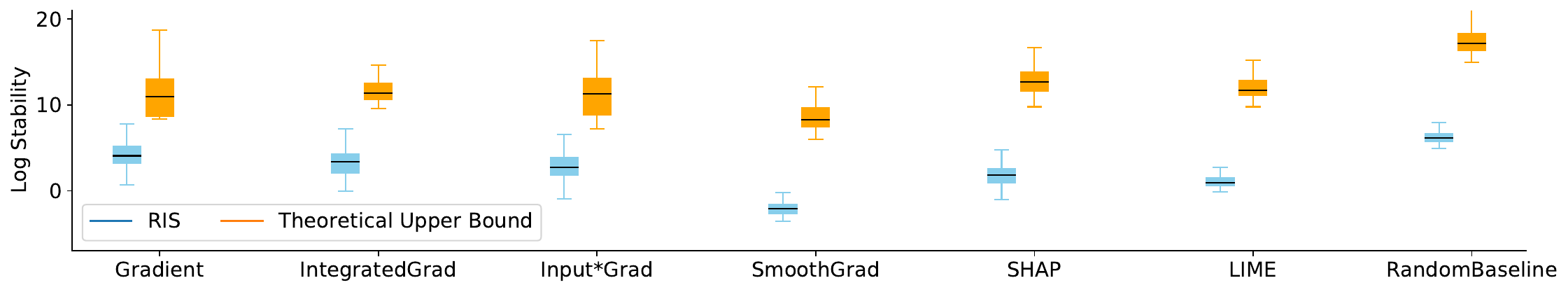}
\end{subfigure}
\vspace{-2mm}
\caption{Theoretical upper bounds for the (log) relative input stability (RIS) computed using the right-hand-side of Equation~\ref{eq:proof_1} across seven explanation methods for an ANN predictor trained on Adult dataset. Results show that RIS is upper bounded by the product of $L_{1}$ and RRS (relative representation stability), where $L_1$ is the Lipschitz constant between the input and hidden layer of the ANN model. Results for the Compas and German dataset are shown in Appendix~\ref{app:bounds_ann}.}
\label{fig:bounds_ann}
\vspace{-2mm}
\end{figure*}

\subsection{Results}
\label{sec:result}
\xhdr{Empirically verifying our theoretical bound} We empirically evaluated our theoretical bounds by computing the LHS of Equation~\ref{eq:proof_1} for all seven explanation methods. Results in Figure~\ref{fig:bounds_ann} illustrate the empirical and theoretical bounds for the Relative Input Stability, confirming that none of our theoretical bounds are violated. In addition, we observe that, on average across all explanation methods, our upper bounds are tight with the mean theoretical bounds being 233\% higher than that of the empirical values. Similar results are found for other datasets in Appendix~\ref{app:bounds_ann}.
\begin{figure*}[t]
    \centering
    \vspace{-2mm}
    \begin{subfigure}{0.95\linewidth}
        \centering
        \includegraphics[width=0.95\textwidth]{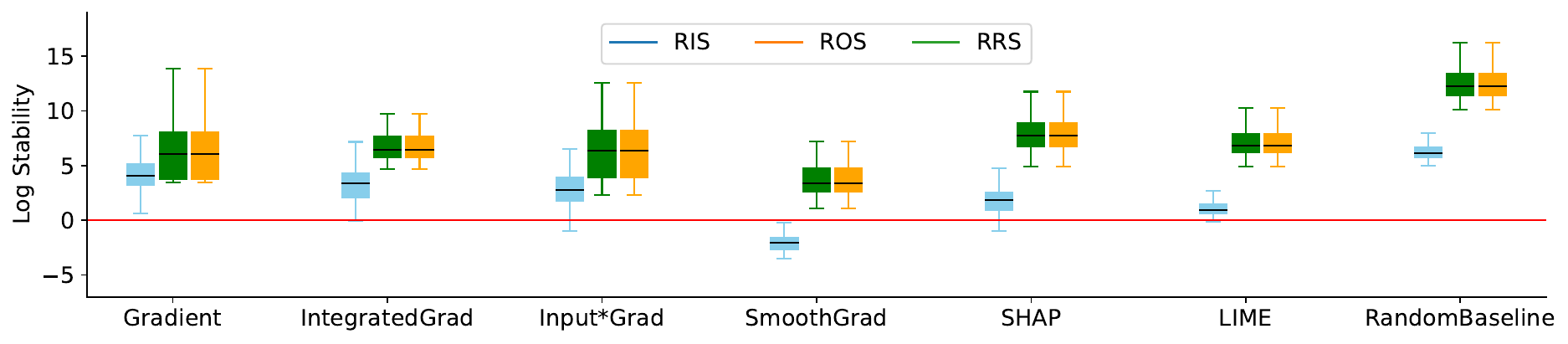}
        \vspace{-2mm}
        \caption{Adult dataset}
    \end{subfigure}
    \begin{subfigure}{0.95\linewidth}
        \centering
        \includegraphics[width=0.95\textwidth]{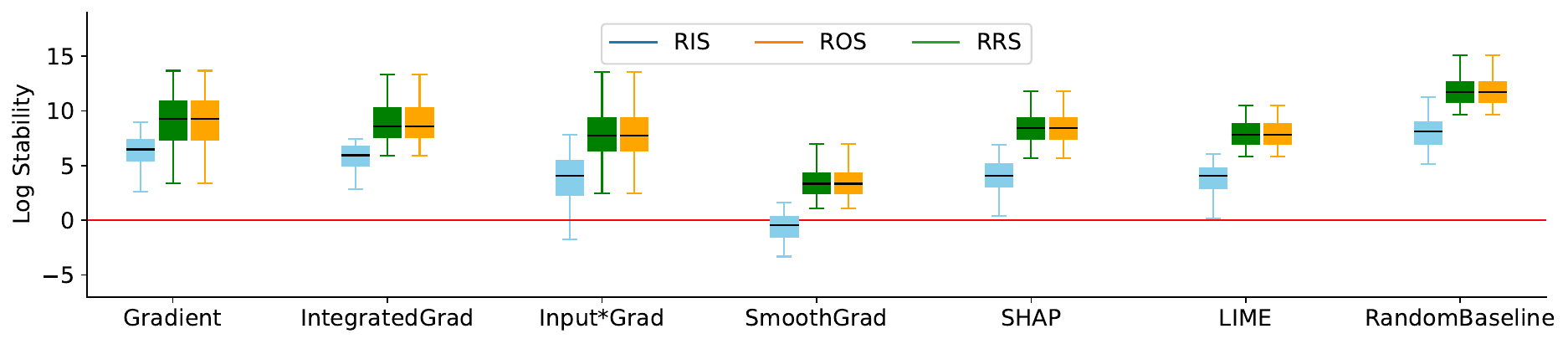}
        \vspace{-2mm}
        \caption{Compas dataset}
    \end{subfigure}
    \begin{subfigure}{0.95\linewidth}
        \centering
        \includegraphics[width=0.95\textwidth]{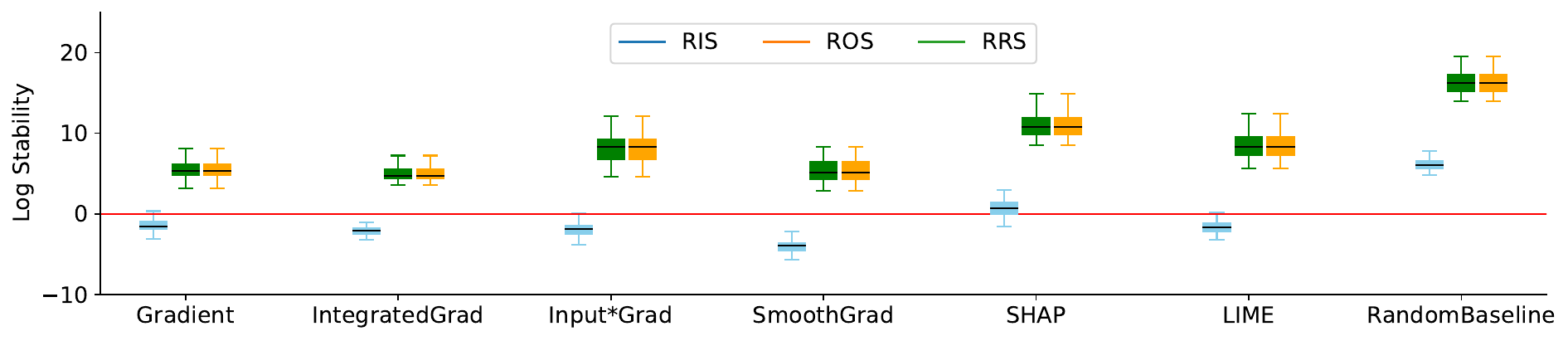}
        \vspace{-2mm}        
        \caption{German Credit dataset}
    \end{subfigure}
    \vspace{-2mm}
    \caption{Empirically calculated log stability of relative stability variants~(Equations~\ref{eq:ris}-\ref{eq:ros}) across seven explanation methods. Results on the Adult (a), Compas (b), and German (c) dataset trained with ANN predictor show that SmoothGrad generates the most stable explanation across RRS and ROS variants. Results for all datasets trained on Logistic Regression models are shown in Appendix~\ref{fig:lr}.}
    \label{fig:ann}
    \vspace{-2mm}
\end{figure*}

\xhdr{Evaluating the stability of explanation methods} We compare the stability of explanation methods by computing instability using all three variants as described in Section~\ref{sec:rel-stab}. Results in Figure~\ref{fig:ann} show that the median instability of all explanation methods using \textit{Relative Input Stability} (Figure~\ref{fig:ann}; blue) are lower than that for the \textit{Representation}~(Figure~\ref{fig:ann}; green) and \textit{Output} Stability~(Figure~\ref{fig:ann}; orange) because the relative input stability~(Equation~\ref{eq:ris}) scores are highly influenced by the input differences ($\bx-\bx'$), i.e., the median RIS scores across all explanation methods are always lower than RRS and ROS. 
Finally, we observe that while no explanation method is completely stable, on average across all datasets and representation stability variants, the SmoothGrad explanation method generates the most stable explanation and outperforms other methods by 12.7\%.


%% file: 050concl.tex
We introduce \textit{Relative Stability} metrics that measure the change in output explanation with respect to the behavior of the underlying predictive model. To this end, we analyze the stability performance of seven state-of-the-art explanation methods using multiple real-world datasets and predictive models. Our theoretical and empirical analysis demonstrate that representation and output stability indicates that SmoothGrad explanation method generates the most stable explanation. We believe that our work is an important step towards developing a broader set of evaluation metrics that incorporate the behavior of the underlying prediction model.

%% file: 060appendix.tex
\section{Theoretical Interpretation}
\label{sec:theory}

Prior works have shown that commonly used artificial neural network (ANN) models comprise of linear and activation layers which satisfy Lipschitz continuity~\citep{gouk2021regularisation}. Let us consider a 2-layer ANN model $f$, where $h_{1}(\cdot)$ and $h_{2}(\cdot)$ represent the outputs of the first and second hidden layers, respectively. For a given input $\bx$ and its perturbed counterpart $\bx'$, we can write the Lipschitz form for the first hidden layer as:
\begin{align}
    ||~h_1(\bx) - h_1(\bx')~||_{p} \leq L_{1}~||~\bx - \bx'~||_{p},
    \label{eq:lipschitz}
\end{align}
where $L$ is the Lipschitz constant of the hidden layer $h_{1}(\cdot)$. Taking the reciprocal of Equation~\ref{eq:lipschitz}, we get:
\begin{align}
    \frac{1}{||~h_1(\bx) - h_1(\bx')~||_{p}} > \frac{1}{L_{1}}~\frac{1}{||~\bx - \bx'~||_{p}},
\end{align}
Multiplying both sides with $||~\frac{(\mathbf{e}_{\bx}{-}\mathbf{e}_{\bx'})}{\mathbf{e}_{\bx}}||_p$, we get:
\begin{align}
\frac{||~\frac{(\mathbf{e}_{\bx}{-}\mathbf{e}_{\bx'})}{\mathbf{e}_{\bx}}||_p}{||~h_1(\bx) - h_1(\bx')~||_{p}} > \frac{1}{L_{1}}~\frac{||~\frac{(\mathbf{e}_{\bx}{-}\mathbf{e}_{\bx'})}{\mathbf{e}_{\bx}}||_p}{||~\bx - \bx'~||_{p}},
\label{eq:rel_lipschitz}
\end{align}
With further simplifications, we get:
\begin{align}
    \frac{||~\frac{(\mathbf{e}_{\bx}{-}\mathbf{e}_{\bx'})}{\mathbf{e}_{\bx}}||_p}{||h_1(\bx)||_{p}||\frac{h_1(\bx) - h_1(\bx')}{h_1(\bx)}||_{p}} > \frac{1}{L_{1}}~\frac{||~\frac{(\mathbf{e}_{\bx}{-}\mathbf{e}_{\bx'})}{\mathbf{e}_{\bx}}||_p}{||\bx||_{p}||\frac{\bx - \bx'}{\bx}||_{p}}
\end{align}
For a given $\bx'$ and representations from model layer $h_{1}(\cdot)$, using Equations~\ref{eq:ris}-\ref{eq:rrs}, we get:
\begin{align}
\frac{\text{RRS}(\bx, \bx', \mathbf{e}_{\bx}, \mathbf{e}_{\bx'})}{||h_1(\bx)||_{p}} > \frac{1}{L_{1}}~\frac{\text{RIS}(\bx, \bx', \mathbf{e}_{\bx}, \mathbf{e}_{\bx'})}{||\bx||_{p}},\\
\Rightarrow \text{RIS}(\bx, \bx', \mathbf{e}_{\bx}, \mathbf{e}_{\bx'}) < \big(L_{1}\frac{||h_1(\bx)||_{p}}{||\bx||_{p}}\big)~\times~\text{RRS}(\bx, \bx', \mathbf{e}_{\bx}, \mathbf{e}_{\bx'}),
\end{align}
where we find that the Relative Input Stability score is upper bounded by $L_{1}$ times $\lambda_{1}{=}\frac{||h_1(\bx)||_{p}}{||\bx||_{p}}$ times the Relative Representation Stability score. Finally, we can also extend the above analysis by substituting $h_1(\cdot)$ with the output logit layer $h_{2}(\cdot)$ and show that the same relation holds for Relative Output Stability:
\begin{align}
\text{RRS}(\bx, \bx', \mathbf{e}_{\bx}, \mathbf{e}_{\bx'}) < \lambda_{2}L_{2}~\times~\text{ROS}(\bx, \bx', \mathbf{e}_{\bx}, \mathbf{e}_{\bx'}),
\end{align}
where $\lambda_{2}=||h_1(\bx)||_{p}$.

\section{Implementation Details}\label{apdx:impl}
\xhdr{Predictors} We train logistic regression (LR) and artificial neural network (ANN) models. Details in Appendix \ref{apdx:model-training}. 
The ANN models have 1 hidden layer of width $100$ followed by a ReLU activation function and the output Softmax layer.

\xhdr{Predictor Training}
\label{apdx:model-training}
To train all predictive models, we used a 80-10-10 train-test-validation split.  We used the RMSProp optimizer with learning rate $2e-03$, the binary cross entropy loss function, and batchsize $32$.  We trained for $100$ epochs and selected the model at the epoch with the highest validation set accuracy as the final prediction model to be explained in our experiments.  

\xhdr{Explanation Method Implementations}
\label{apdx:impl-exp-methods}
We used existing public implementations of all explanation methods in our experiments.  We used the following \texttt{captum} software package classes: i) \texttt{captum.attr.Saliency} for VanillaGrad; ii) \texttt{captum.attr.IntegratedGradients} for IntegratedGradients; iii) \texttt{captum.attr.NoiseTunnel}; iv) \texttt{captum.attr.Saliency} for SmoothGrad; v) \texttt{captum.attr.InputXGradient} for Gradients$\times$Input; and vi) \texttt{captum.attr.KernelShap} for SHAP. We use the authors' LIME \href{https://github.com/marcotcr/lime}{python package} for LIME.  

\xhdr{Metric hyperparameters} For all metrics, we generate a neighborhood $\mathcal{N}_{\bx}$ of size $50$ for each point $\bx$. The neighborhood points were generated by perturbing the clean sample $\bx$ with noise from $\mathcal{N}(\bx, 0.05)$. For data sets with with discrete binary inputs we used independent Bernoulli random variables for the pertubations: for each discrete dimension, we replaced the original values with those that were drawn from a Bernoulli distribution with parameter $p=0.03$. Choosing a small $p$ ensures that only a small fraction of samples are perturbed to reduce the likelihood of sampling an out-of-distribution point. For internal model representations $\cL_{\bx}$ we use the pre-softmax input linear layer output embedding for the LR models, and the pre-ReLU output embedding of the first hidden layer for the ANN.

\input{table-explanation-hyperparams}

\begin{figure}[t]
    \centering
    \vspace{-2mm}
	\begin{subfigure}{0.95\linewidth}
    	\centering
    	\includegraphics[width=0.95\textwidth]{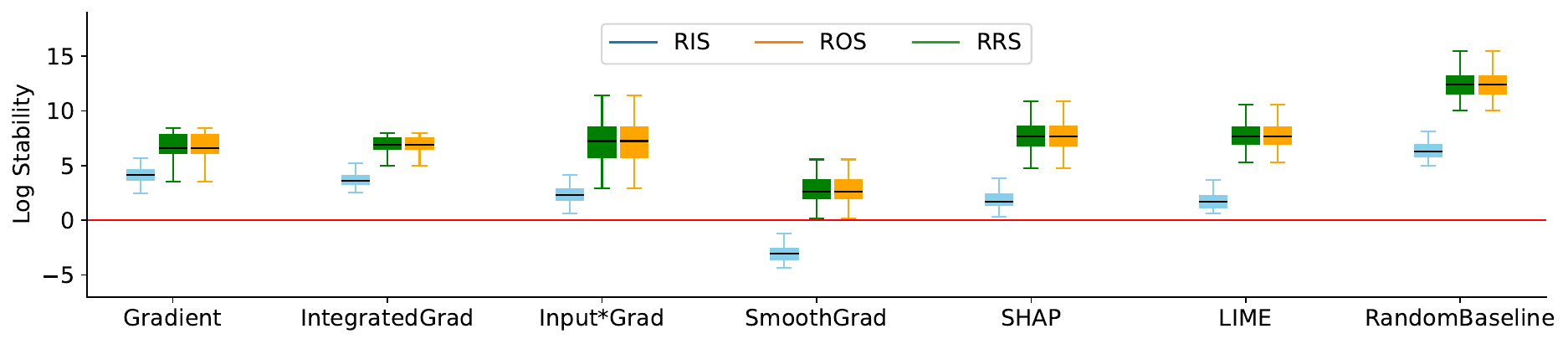}
	\end{subfigure}\\
	\small\vspace{0.5mm}(a) Adult dataset\\
	\begin{subfigure}{0.95\linewidth}
    	\centering
    	\includegraphics[width=0.95\textwidth]{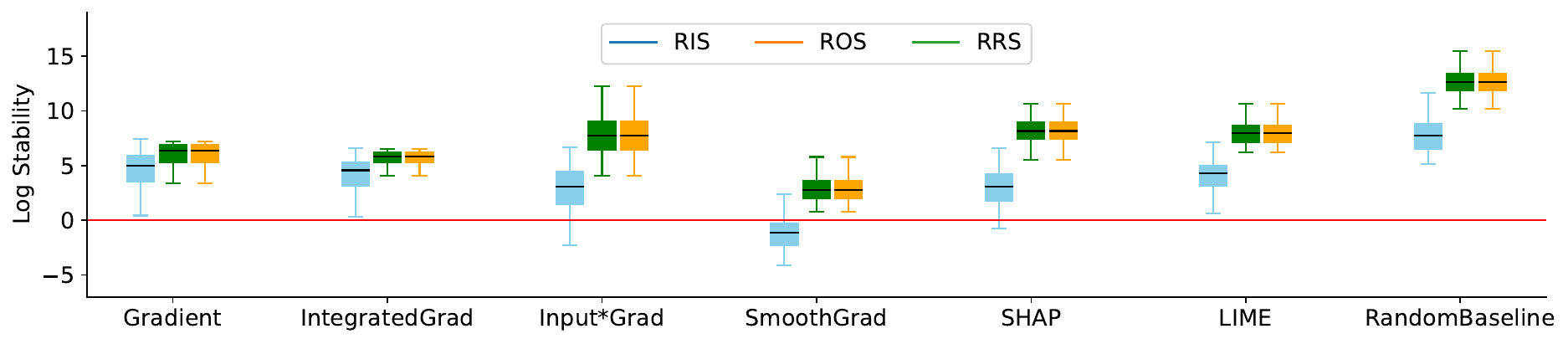}
	\end{subfigure}\\
	\small\vspace{0.5mm}(b) Compas dataset\\
	\begin{subfigure}{0.95\linewidth}
    	\centering
    	\includegraphics[width=0.95\textwidth]{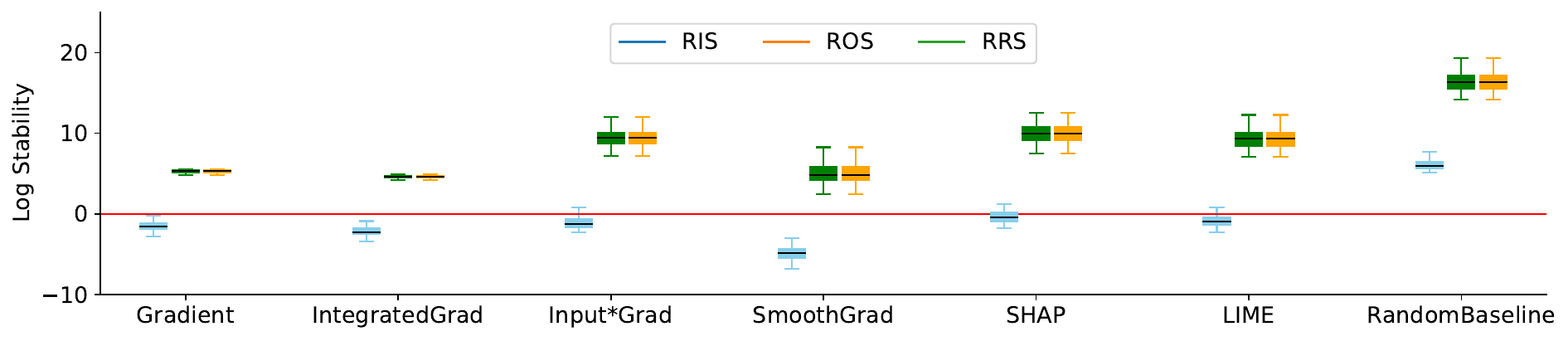}
	\end{subfigure}\\
	\small\vspace{0.5mm}(c) German dataset\\
	\caption{Empirically calculated log stability of all three relative stability variants~(Equations~\ref{eq:ris}-\ref{eq:ros}) across seven explanation methods. Results on the Adult dataset trained with Logistic Regression predictor show that SmoothGrad generates the most stable explanation across representation and output stability variants.}
    \label{fig:lr}
\end{figure}

\begin{figure*}[ht]
\centering
\vspace{-2mm}
\begin{subfigure}{0.95\linewidth}
\centering
\includegraphics[width=0.95\textwidth]{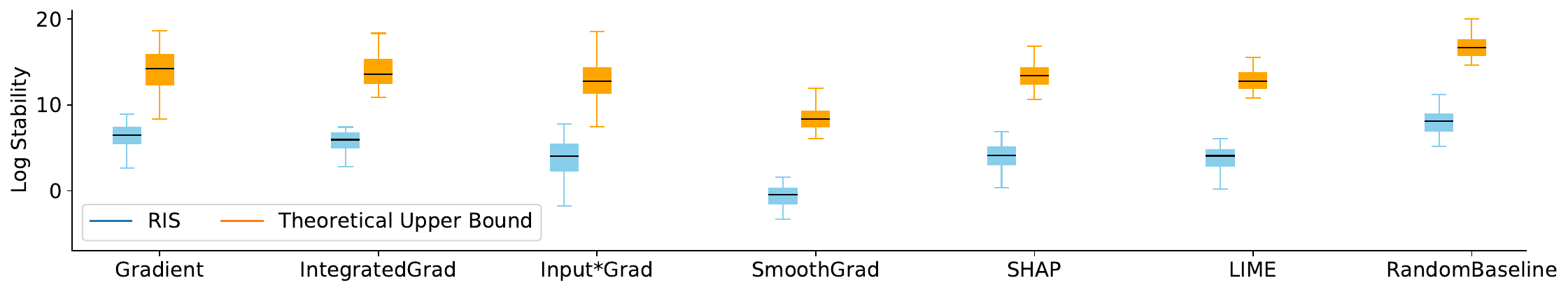}
\caption{Compas dataset}
\end{subfigure}
\vfill
\begin{subfigure}{0.95\linewidth}
\centering
\includegraphics[width=0.95\textwidth]{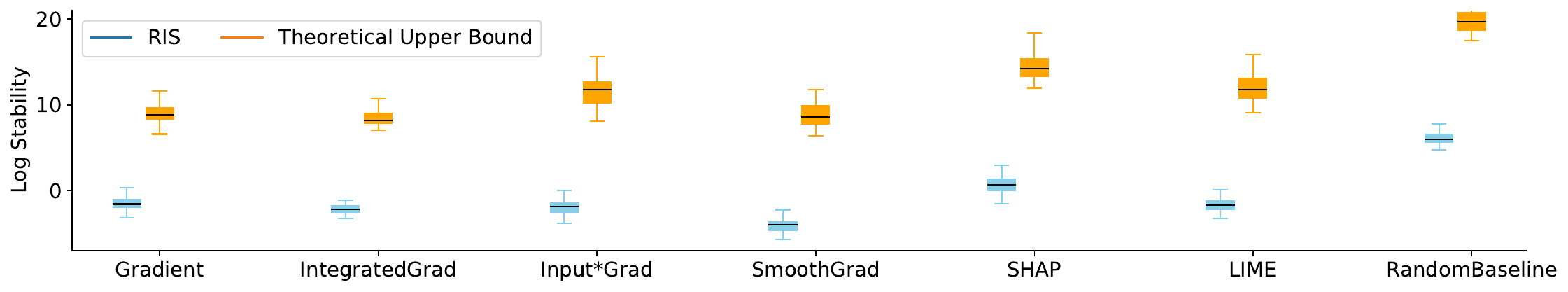}
\caption{German credit dataset}
\end{subfigure}
\caption{Theoretical upper bounds for the (log) relative input stability (RIS) computed using the right-hand-side of Equation~\ref{eq:proof_1} across seven explanation methods for an ANN predictor trained on the Compas and German credit datasets. Results show that RIS is upper bounded by the product of $L_{1}$ and RRS (relative representation stability), where $L_1$ is the Lipschitz constant between the input and hidden layer of the ANN model.}
\label{app:bounds_ann}
\end{figure*}


%% file: table-explanation-hyperparams.tex
\begin{table}[ht!]
\centering\small
\renewcommand{\arraystretch}{0.8}
\setlength{\tabcolsep}{1.2pt}
\begin{tabular}{lcc} 
Explanation Method & Hyperparameter & Value\\
\toprule
 & \texttt{n\_samples} & $1000$ \\
LIME & \texttt{kernel\_width} & $0.75$ \\
 & \texttt{std} & $0.05$ \\
\midrule
SHAP & \texttt{n\_samples} & $500$  \\
\midrule
SmoothGrad & \texttt{std} &  0.05 \\
\midrule
Integrated Gradients & \texttt{baseline} & train data means \\
\midrule
Random Baseline & attributions from $\mathcal{N}(0,1)$ \\
\bottomrule
\end{tabular}
\caption{Hyperparameters used for explanation methods. For hyperparameters not listed, we used their package defaults.}
\end{table}